\def\year{2021}\relax
\documentclass[letterpaper]{article} % DO NOT CHANGE THIS
\usepackage{aaai21}  % DO NOT CHANGE THIS
\usepackage{times}  % DO NOT CHANGE THIS
\usepackage{helvet} % DO NOT CHANGE THIS
\usepackage{courier}  % DO NOT CHANGE THIS
\usepackage[hyphens]{url}  % DO NOT CHANGE THIS
\usepackage{graphicx} % DO NOT CHANGE THIS
\usepackage{graphicx}  % DO NOT CHANGE THIS
\usepackage{natbib}  % DO NOT CHANGE THIS AND DO NOT ADD ANY OPTIONS TO IT
\usepackage{caption} % DO NOT CHANGE THIS AND DO NOT ADD ANY OPTIONS TO IT
\usepackage{latexsym}
\usepackage{xcolor}
\usepackage{amsmath}
\usepackage{algpseudocode}
\usepackage{algorithm,algpseudocode}
\usepackage{amsfonts}
\usepackage{amssymb}
\usepackage{xspace}
\usepackage{comment}

\usepackage[switch]{lineno}

\newcommand{\vToq}{vision-to-question\xspace}

\title{Let's Talk! Striking Up Conversations via\\Conversational Visual Question Generation}
\author{
\begin{tabular}[c]{@{}c@{}}
    Shih-Han Chan$^{1}$~~~~
    Tsai-Lun Yang$^{1}$~~~~
    Yun-Wei Chu$^{1}$~~~~
    Chi-Yang Hsu$^{1, 2}$\\
    Ting-Hao (Kenneth) Huang$^{2}$~~~~
    Yu-Shian Chiu$^{3}$~~~~
    Lun-Wei Ku$^{1}$
    \end{tabular}\\
    $^{1}$\normalfont Academia Sinica, 
    $^{2}$\normalfont The Pennsylvania State University, 
    $^{3}$\normalfont Institute for Information Industry\\
    \normalfont $^{1}$\{hank08tw,a7532ariel,yunweichu,lwku\}@iis.sinica.edu.tw,
    \normalfont $^{2}$\{cxh5437,txh710\}@psu.edu, $^{3}$samuelchiu@iii.org.tw
}    

\begin{document}

\maketitle

%\linenumbers
\begin{abstract}

An engaging and provocative question can open up a great conversation. In this work, we explore a novel scenario:
a conversation agent views a set of the user's photos (for example, from social media platforms) and asks an engaging question to initiate a conversation with the user.
The existing \vToq models mostly generate tedious and obvious questions, which might not be ideals conversation starters.
This paper introduces a two-phase framework that first generates a visual story for the photo set and then uses the story to produce an interesting question. The human evaluation shows that our framework generates more response-provoking questions for starting conversations than other \vToq baselines.

\end{abstract}

%\kenneth{Do NOT use the phrase ``responsive questions'' **IN THE WHOLE PAPER**. This word doesn't make sense. Use ``provocative questions'' or ``response-provoking questions'' instead.}
%\yun{revised}

\section{Introduction}

%懷舊治療Reminiscence Therapy

%motivate
Question-asking play an essential role in human conversations.
%An engaging and provocative question can open up a great conversation.
%Conversation is an essential human behavior.
%People can contribute to an ongoing conversation in many ways, such as 
%talking, question-asking, and replying, it requires someone to open.
Studies have shown that people who ask more questions in interpersonal conversations are better liked by their conversational partners~\cite{huang2017doesn}.
%gao-etal-2019-interconnected
%Asking a question is the most effective way to initiate a conversation, because it encourages another person to answer
For automated social bots, prompting the user with a question is known to be an effective way to initiate conversations.
%\kenneth{FIND CITATIONS. Try to add 1 or 2 sentences to describe each paper you added.}\yun{added}.
For example,~\citet{wang-etal-2018-learning-ask} generates questions in diverse yet relevant topics to enhance the interactiveness and persistence of conversations.
\citet{pan-etal-2019-reinforced} also uses a Reinforced Dynamic Reasoning network to produce meaningful questions to engage users in conversations.
%scenario
In this work, we explore a novel scenario: an automated conversation agent ``views'' a user's photos --- for example, from social media platforms, or shared by the user proactively --- and asks an engaging question to initiate a conversation with the user.
This scenario uses images shared by the user to start a conversation, enriching the plausible topics of the human-agent conversations.
%allowing for broader and more rich topics.
For example, when a user posts a set of wedding photos with only a little or even no descriptions, the proposed conversational agent can still ask questions about the wedding.
Such machine-generated questions can be used to engage user for social purposes;
to allow robots to elicit situational information from passersby~\cite{krishna2019information}; 
or to support memory therapies ({\em e.g.}, reminiscence therapy) or activities that use the patients personal images as memory prompts~\cite{bhar2014reminiscence}.

%Furthermore, generating human-centered questions based on images has several applications, 
%has several real-world applications
%Asking a question from an image has many applications in several fields such as helping visually-impaired and Reminiscence Therapy~\cite{bhar2014reminiscence}.
%Generating questions that starts conversation can also be applied to part of Reminiscence Therapy. 
%Reminiscence Therapy is a treatment that uses all the senses (e.g., sight, touch, taste, etc.) to help individuals with dementia remember events, people and places from their past lives. 
%As part of the therapy, if a model can take images that contain some individual's past information as input and generate a question based on images to start a conversation, this will bring memories from the distant past into present awareness and help the treatment.

\begin{figure}[t]
     \centering
     \includegraphics[width=\linewidth]{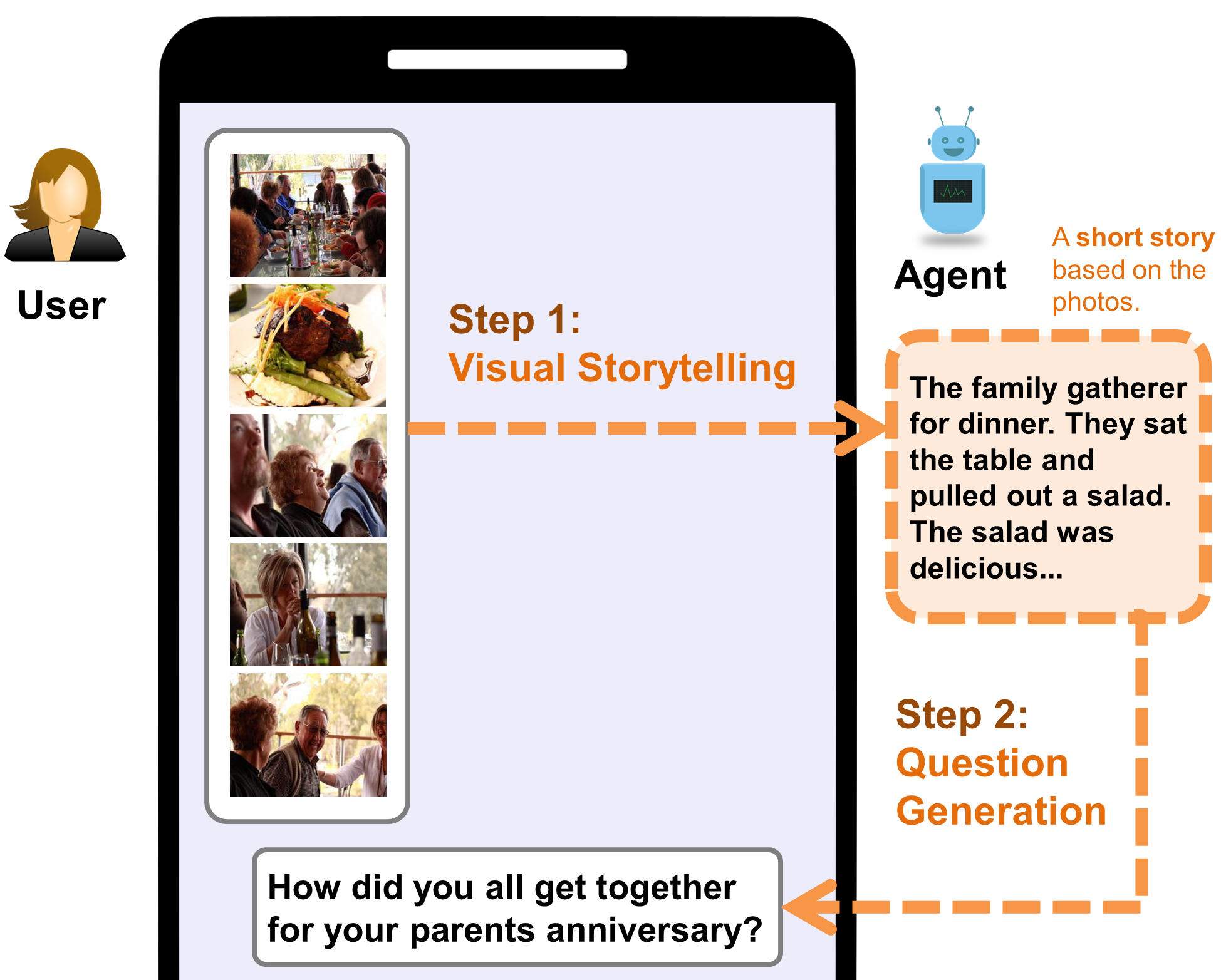}
     \caption{The agent receives five images from the user and comes up with an short story based on the images. Then the agent asks a response-provoking question based on the story.}
\label{fig:scenario}
\end{figure}

%that were not mentioned in the user's textual posts or previous conversations.
%elicit questions and thus suitable to start social conversations 
%rather than to produce questions based on prior conversations.
%initiating a conversation cab be challenging. 
%\kenneth{---------------- KENNETH IS STILL WORKING HERE ------------------}
%Most of the text-based question generation models for conversational usage lack material to initiate a conversation, because this kind of models usually generates questions based on previous historical conversation.
%By bringing external sources ({\em e.g.}, images), previous work introduced a vision-to-question task that asks a question based on a given image.
However, the existing \vToq 
%\kenneth{Let's use ``\vToq'' rather than vision-to-question. Please update the whole paper using the command ``\vToq''.} 
models, such as visual question answering (VQA)~\cite{DBLP:journals/corr/AntolALMBZP15}, do not generate questions with the purpose of engaging users or provoking users' responses~\cite{krishna2019information}.
For example, many VQA questions are about the factual property, such as color, size, and shape, of the objects in the image, rather than human activities or broader contexts of the image.
%\kenneth{1. is this even true? 2. any references?} \cite{DBLP:journals/corr/ZhuGBF15}\yun{cited}.
%When generating a question from a given image, traditional vision-to-question work usually generates questions that ask what is in the photo.
%However, this kind of methods failed to yield a more responsive question that invokes humans' desire to chat.
In this paper, %we propose a unique scenario that is ignored by previous work.
instead of producing generic questions using one image, 
our proposed model takes a sequence of images as input and generates an engaging and provocative question based on these images to start a conversation with the user.

%focus on generating generic questions
%mostly generate tedious and obvious questions, which might not be ideals conversation starters.
%solution
This paper introduces a two-phase framework that first generates a \textbf{visual story} for the image sequence and then uses the story to produce an interesting question.
This approach takes advantage of the existing ``visual storytelling'' (VIST) technologies, where the model generates an engaging short story based on a sequence of images~\cite{huang2016visual}.
We use one of the state-of-the-art visual storytelling model, KG-Story~\cite{hsu2020knowledge}, to produce a visual story, which is then fed into a Transformer-based model to create engaging questions.
%Therefore, if a system can first develop an attractive story that describes the scenario of images then generates a question based on the story, the question might be more responsive and engaging.
%Hence, we propose a framework that the system first performs story generation and then asks an engaging question based on the imagining stories.
Figure~\ref{fig:scenario} illustrates the scenario of the proposed framework.
The human evaluation shows that our framework generates more response-provoking questions than other \vToq baselines.

The contribution of this work is three-fold: 
\begin{itemize}

\item We are the first to introduce a unique scenario: inputting a sequence of images, the system then asks the user an engaging and response-provoking question to start a conversation.

\item We propose a two-stage framework to perform visual question generation: the system first generates stories based on the image sequence. It then generates an engaging question based on the story.

\item We conduct a human evaluation using the VIST dataset to show that the questions generated by the proposed approach are better at invoking human desire to chat than traditional \vToq methods.

\end{itemize}

%------------------- dead kitten ----------------

\begin{comment}
    
\end{comment}

\section{Related Work} 

% \yun{Most related concept is VQG, however no conception.}
% \yun{use caption as conception? but too caption-like-Q}
% \yun{use story}
% \hank{text to question vs image to question}
% Text-based question generation model usually generate questions based on given paragraph or historical conversation.
% some LSTM model...
% With the benefit brought by Transformer for language modeling, many approaches develop question generation model based on Transformer. 
\paragraph{Question Generation}
Question generation usually takes the datasets designed for question answering and generates questions based on given textual context ({\em e.g.}, paragraph or historical conversation).
Most traditional approaches design an end-to-end structure by implementing the recurrent neural network~\cite{duan-etal-2017-question}.
Some implement attention mechanism~\cite{bahdanau2014neural} to enhance embedding features~\cite{DBLP:journals/corr/ZhouYWTBZ17}.
With the benefit brought by Transformer~\cite{devlin2018bert} for language modeling, \citet{QGbyTransformer} starts developing question generation model based on Transformer. 
However, existing text-based question generation is hard to start a conversation since the lack of informative material, and most works still use historical conversation to generate a question.
% However, it's hard to implement text-based question generation on robot for starting a conversation, since robot lacks textual input.

% There are several vision-to-language tasks that related to proposed framework.
% We introduce visual question generation, image captioning, and visual storytelling in the following paragraph.

\paragraph{Visual Question Generation}
With the help of external sources, Visual Question Generation (VQG) aims to generate a question based on a given image.
The idea of VQG came from Visual Question Answering (VQA)~\cite{DBLP:journals/corr/AntolALMBZP15}. 
However, the questions in the VQA task are designed limited to objects, colors, numbers, or locations. \citet{DBLP:journals/corr/MostafazadehMDZ16} introduced the VQG dataset, where the system is asked to generate a question for people to answer. 
% , not just focus on the ability of computer vision.
Most of approaches focus on leveraging seq2seq model to generate questions~\cite{DBLP:journals/corr/abs-1808-03986}.
% \citet{DBLP:journals/corr/abs-1808-03986} leveraged seq2seq model to generate questions.
% Some approaches implement Variational Autoencoder (VAE) to generate multiple and diverse types of questions for a given image\citet{DBLP:journals/corr/JainZS17,krishna2019information}.
% \citet{DBLP:journals/corr/JainZS17} proposed a generative method using Variational Autoencoder (VAE), which can generate multiple and diverse types of questions for a given image.
% Another VAE-based approach \cite{krishna2019information} aims to generate questions that are tight to a given category. 
% In our work, we also want our system to generate an interesting question.
% Furthermore, we want the question to make people willing to keep the conversation continue and not just stop after answering.
However, generating questions from the image feature lacks a comprehensive understanding of visual input as the model still briefly asks questions about the objects in the image.
% \yun{ending of this paragraph: some cons of VQG(still asking boring question). However, generating question from image feature lack comprehensive understanding of visual input. We propose let the model first understand what happen in the image then asking a question.}

\paragraph{Image Captioning}
We conducted a survey on image captioning, a task that the model should use a sentence to describe one image, which can help comprehensively understand the visual inputs.
The model should learn representations of the interdependence between the objects/concepts in the image and use them to describe the image factually. 
% Template-based approaches have fixed templates with a number of blank slots to generate captions \cite{10.1007/978-3-642-15561-1_2, 5995466}, while retrieval-based methods select captions from a candidate pool \cite{NIPS2011_4470}.
\citet{pmlr-v32-kiros14} first introduced a deep learning based method on this topic by using CNN to extract image features and a language model to generate captions.
Some work proposed architecture or applied attention mechanism based on recurrent neural network~\cite{DBLP:journals/corr/VinyalsTBE14, DBLP:journals/corr/XuBKCCSZB15}.
However, describing the image is usually not considered attractive by humans. If we want our system to communicate with humans, it must capture their interests and avoid just stating the obvious.

\paragraph{Visual Storytelling}
%1. 先講vqg(inspired from VQA)產生的問題有什麼缺點（不適合做為起使對話，不有趣．因為適用vqa生成的，通常是問資訊ex.有幾顆頻果） 2. captioning 通常是匯總圖片中的資訊跟重點，無法生成有趣的內容 3. story 通過object detection, term prediction, knowledge graph生成的story 有豐富的內容還有不同東西間的關係，加上一點想像跟聯想，能生成有趣想像力的故事，再通過transformer生成的問題也會比較有趣讓人能作為起始問題
Visual storytelling (VIST) was proposed by \citep{huang2016visual}.
Unlike image captioning which generates a sentence describing the image, VIST asks the model to generate a story based on 5 images.
Visual stories should be descriptive, relevant to the images, and appealing to human readers.
Most of the approaches focus on developing end-to-end models or adopting various training techniques on the VIST dataset~\cite{kim2018glac}.
Since there is only one existing VIST dataset, these end-to-end VIST works often limit the knowledge to this dataset.
To avoid overfitting on the dataset, some research leverage external sources to enrich story contents.
Since knowledge graphs have shown beneficial on language modeling, most of the VIST work use knowledge graphs (KG) to enrich stories~\cite{hsu2020knowledge}.
% Most of VIST work use knowledge graphs (KG) to enrich stories~\cite{Wang2020StorytellingFA, hsu2020knowledge}, since knowledge graphs have shown beneficial on language modeling~\cite{ bowman-etal-2015-large, zhang-etal-2017-ordinal}.
Among all the KG-based VIST, KG-Story~\cite{hsu2020knowledge} designed a three-stage framework, which uses Visual Genome knowledge graph~\cite{Krishna_2017} to add semantic relations between two adjacent images, generated more interesting and coherent stories. 
Since stories can generate more creative content from visual inputs than captions, our framework takes visual stories to produce response-provoking questions.

% \hank{Due to the creativity and coherence of the story, our model generates story as an ingredient to create interactive question.}

% text to question 
% People usually don't know what to talk, so there is no raw text. We trans from image to a interesting story (text) by utilizing knowledge graph, then apply text to question method to generate interactive question. 

% image (social) commenting
% Given an image with description, image commenting model can generate funny and attractive question (sentence) based on subject in the description, however, the question (sentence) generated is not suitable for starting a conversation with human, since conversation requires interactive communication.

\section{Methods}

% \begin{figure*}[t]
%      \centering
%      \includegraphics[width=\linewidt{images/flowchart1.png}
%      \caption{The overall architecture of Story question generation model}
% \label{fig:scenario}
%  \end{figure*}
\begin{figure*}[t]
     \centering
     \includegraphics[width=\linewidth]{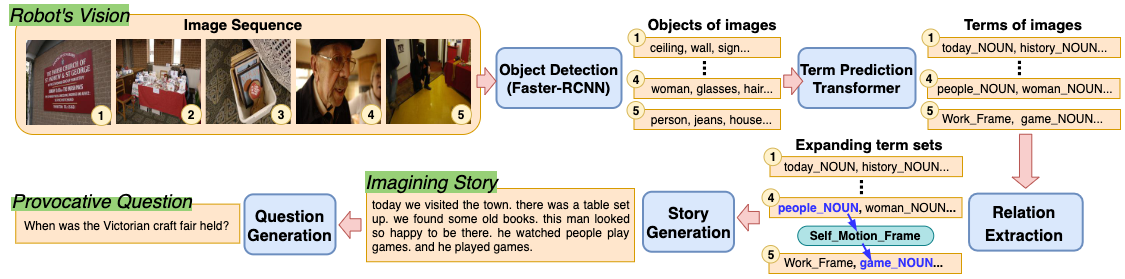}
     %\vspace{-1.5pc}
	  \caption{The pipeline of proposed framework. We first ask the system to think by generating stories based on given image sequence. The system then produces provocative question based on the story.
	  }
     %\vspace{-1pc}
\label{fig:overview}
 \end{figure*}

The overall framework contains two stages, as described in Figure~\ref{fig:overview}. 
The system first generates stories based on input vision as intermediate perception.
From the generated stories, the system then performs question generation and outputs an response-provoking question.

\subsection{Stage 1: Visual Storytelling (KG-Story)}

We implement KG-Story\footnote{KG-Story: https://github.com/zychen423/KE-VIST}~\cite{hsu2020knowledge}, a three-stage story generation framework, as our story generation model.
KG-Story extracts representative terms from input images, 
enriches terms by knowledge graph, and eventually generates stories based on enriched terms. 
%In the following, we briefly describe each step.
%Please read their papers for more details.

\begin{enumerate}
    \item \textbf{Extracting representative terms:}
Given a sequence of images, KG-Story uses a pre-trained Faster-RCNN~\cite{ren2015faster} as the object detection model. 
To reduce computational complexity, only the objects within the top 25 confidence scores are used.
Since objects lack semantic meaning, KG-Story designs a Transformer based model that transforms objects to terms ({\em e.g.}, objects and actions).
From VIST dataset, they use SpaCy\footnote{SpaCy: https://spacy.io/} and Open-SASEME~\cite{swayamdipta17open} to parse stories into object nouns and semantic frames as ground truth terms.
Taking predicted objects from Faster-RCNN as input and parsed terms as output, a Transformer encoder ~\cite{vaswani2017attention} and a GRU decoder with an attention mechanism ~\cite{bahdanau2014neural} are used as term prediction model.

\item  \textbf{Knowledge enrichment:}
Since previous end-to-end models tend to generate caption-like incoherent stories which are relatively boring, KG-Story enriches stories by introducing knowledge graph. 
Knowledge graph serves as the source of ideas that connects two images and ensures the coherence of logic.
KG-Story link terms in two adjacent images using the relations provided by Visual Genome knowledge graph~\cite{Krishna_2017}.

Given 5 images, the extracted terms from Stage 1 are represented as $\{m^1_1,...,m^t_i,...,m^5_{N_5}\}$, where $\{m^1_1,...,m^1_h\}$ denotes first image's term set,  $m^t_i$ denotes the $i$-th term from image $t$ and $N_k$ is the number of terms from image $k$.
From consecutive images, KG-Story explores all possible relations $(m^t_i, r, m^{t+1}_j)$ and  $(m^t_i, r_1, m_{\mathit{middle}}, r_2, m^{t+1}_j)$, while $m_{\mathit{middle}}$ denotes a knowledge graph entity that bridges $m^t_i$ and $m^{t+1}_j$.
With all possible relations, KG-Story uses a RNN-based language model to obtain a relation with lowest perplexity.
The chosen relation is inserted to original term sequence expanding the number of term sets from 5 to 6.

\item \textbf{Story generation:}
To generate stories, KG-Story leverages Transformer~\cite{vaswani2017attention} with expanded term sets from Stage 2 as input. 
Three modifications are made for the original Transformer model.
(1) The length-difference positional encoding is adopted to perform variable-length story generation. 
Since all the samples of VIST dataset contain five images, this mechanism allows KG-Story to generate additional sentence.
(2) Anaphoric expression generation is used for the unification of anaphor representation.
To enable the use of pronouns, KG-Story uses a coreference resolution tool \footnote{NeuralCoref 4.0: Coreference Resolution in spaCy with Neural Networks. https://github.com/huggingface/neuralcoref} on the stories to find the original mention of each pronoun.
(3) A designed repetition penalty for inter- and intra-sentence with beam search are adopted to reduce redundancy.
After feeding term sets into designed Transformer, the model generates a knowledgeable story with 6 sentences.
We then use stories as robot's perception and generate an response-provoking question in the next step.
\end{enumerate}

\subsection{Stage 2: Response-Provoking Question Generation from the Story}
We utilize a Transformer-based end-to-end model~\cite{lopez2020transformerbased} as our question generation model.
For the pre-trained model, we use Hugging-Faces implementation~\cite{wolf2020huggingfaces} of the 60 million parameters T5, the smallest of the five available T5 model sizes.

As pre-trained T5 has shown a strong ability to solve the text-to-text problem, we finetune it by taking the paragraphs of Stanford Question Answering Dataset (SQuAD) as input and the question as output. %To be specific, 
The entire dataset is firstly transformed into a continuous body of text; each training sample consists of a context paragraph and associated question(s) transformed into a single continuous sequence with a ``delimiter'' in between. During training, the delimiter enables the model to successfully distinguish between context paragraph and corresponding question(s), while during inference, the delimiter acts as a marker at the end of context to invoke question generation behavior of the model.  

After pretraining and finetuning, the model then takes the generated stories as input and generates a question. Higher Temperature values give more randomness to question generation, while lower values approach greedy behavior. We set Temperature to 0.6.

The generation process use the top-p nucleus sampling method with a value $p = 0.9$, which allows for more diverse generations than a
purely greedy scheme, and minimizes the occurrence of certain tokens or token spans repeating indefinitely in the generated text. 
\newcommand*{\escape}[1]{\texttt{\textbackslash#1}}
For each context paragraph input, the question generation stops when the model reaches the generation length of 26 tokens or the model generates a newline character \escape{n}. We set the maximum length to terminate the question generation of some context that don not reach the \escape{n} on their own.

%\kenneth{Do not use ``don't'' in papers. Use ``do not''.} \yun{revised}
% \subsubsection{Input representation}
% \subsubsection{main model}
% \subsubsection{generation}

\section{Experimental Setups}

% We first introduce the dataset we use for model training and performance evaluation.
% Baseline....
% Human evaluation using AMT to rank the performance...
In this section, we first introduce the datasets we use for model training and performance evaluation. We then provide details of the baseline models we compare to and all models’ hyperparameter settings.
Lastly, we crowdsource our human evaluation on Amazon Mechanical Turk: ask the workers to rank the performance of generated questions and conduct user study.
% Lastly, we crowdsource our human evaluation on Amazon Mechanical Turk, asking the workers to rank the performance of generated questions.

\subsection{Data Preparation}
Four datasets are used in this paper: Visual Genome, ROCStories, SQuAD, and VIST.
% , Charades-Ego, and Pepper-life.
For story generation part, VIST is used to extract terms from images (Stage 1) and fine-tune the story generation model (Stage 3).
Visual Genome knowledge graph is used for relation linking (Stage 2) between the extracted terms.
ROCStories supplies a large quantity of pure textual stories for pre-training the story generator (Stage 3). 
For question generation part, Stanford Question Answering Dataset (SQuAD) is used to train the question generation model.
The test data of VIST are used to examine the performance of our proposed framework.
% , Charades-Ego, and Pepper-life dataset are used to examine the performance of our proposed framework.
The detail of each dataset is described bellow.

\begin{itemize}
    \item \textbf{Visual Genome:}
    Visual Genome~\cite{Krishna_2017} is a
knowledge-based dataset that connects images concepts to language.
It has 108,077 images, 3.8 million object instances, and 2.3 million relationships.
The knowledge graph we utilize covers nouns and relations, categorized into semantic frames, provided by the scene graph of Visual Genome.
Compared with most image-to-text works that focus on objects nouns and generate static stories, adding logical relation and activities frames makes our stories reasonable and vivid.

\item \textbf{ROC-Stories Corpora:}
We use the ROC-Stories~\cite{mostafazadeh2016corpus}, which contains 98,159 pure textual stories, to pre-train our story generator.
As the annotators were asked to write five-sentence stories given a prompt,
ROC-Stories focuses on specific, everyday topics. 

\item \textbf{SQuAD:}
SQuAD~\cite{DBLP:journals/corr/RajpurkarZLL16} is a reading comprehension dataset consisting of 100,000+ question-answer pairs posted by crowdworkers on a set of Wikipedia articles, which contain 23,215 paragraphs covering a wide range of topics.
We use the paragraphs and questions in SQuAD to train the question model.

\item \textbf{VIST:}
VIST~\cite{huang2016visual} is a sequential vision-to-language dataset that moves visual understanding from basic perspective to more human-like understanding of grounded event structure.
We train KG-Story model and conduct experiments on VIST, which includes 10,117 Flicker albums with 210,819 unique images.
We follow the standard split setting as previous work, with 40,098 samples for training, 4,988 for validation, and 5,050 for testing. 
Each sample contains one story that describes five images from a photo stream.

\end{itemize}

\subsection{Hyperparameter Configuration}
In all of our experiment, we use the same hyperparameters the authors mentioned in the KG-Story paper.
For term prediction (Stage 1) and story generation (Stage 3), the hidden size is set to 512.
The number of head and layer of the Transformer encoder are 2 and 4.
All KG-Story models are trained with Adam optimizer~\cite{DBLP:journals/corr/KingmaB14} with initial learning rate 1e-3.

For the training parameters of question generation model, the model is trained for 3 epochs using general language modeling loss. 
Adam optimizer was also applied with an initial learning rate of 5e-5 and a linearly decreasing learning rate scheduler with warm up for $10$\% of total training steps.

\subsection{Baselines}
We compare our proposed story-to-question concept with two state-of-the-art frameworks.
The first framework is an end-to-end image question generation model ~\cite{DBLP:journals/corr/MostafazadehMDZ16}, which aims to generate an engaging question given an image. 
The second framework is an image captioning model~\cite{DBLP:journals/corr/XuBKCCSZB15}, which generates a sentence through convolutional neural network and recurrent neural network to describe the content of an image.
We then concatenate captions and perform question generation on  captions.
The purpose is to examine the performance of questions generated by different perception, captions or stories.

\begin{itemize}
    \item \textbf{Image Question Generation:}
We use Gated Recurrent Neural Network\footnote{https://github.com/chingyaoc/VQG-tensorflow} ~\cite{DBLP:journals/corr/MostafazadehMDZ16} as our question generation baseline model.
This model is trained on the VQA dataset \footnote{https://visualqa.org/} ~\cite{DBLP:journals/corr/AntolALMBZP15}, which contains 204,721 COCO images and at least 3 questions per image. The model uses a pre-trained 19-layer VGG Net~\cite{VGG} for encoding image features. It transforms the 4096-dimensional output of the VGG-19’s last fully connected layer (fc7) to a 512-dimensional vector that serves as the initial state of a long-short term memory unit (LSTM) to generate the corresponding question.

\item \textbf{Image Captioning:}
We use the model \citet{DBLP:journals/corr/XuBKCCSZB15} proposed \footnote{https://github.com/sgrvinod/a-PyTorch-Tutorial-to-Image-Captioning} as the image captioning baseline.
The model consists of a 101-layer ResNet pre-trained on the ImageNet classification task, a soft attention network, and a 512-dimensional LSTM. A linear layer is used to map the encoded images to the initial hidden and cell states for the LSTM. The Attention network considers the sequence generated thus far and attends to the part of the image that needs describing next. The LSTM is used to produce the output caption one word at a time conditioned on the context vector, the previous hidden state, and the previously generated word. The whole model is trained on the MS COCO '14 Dataset \footnote{https://cocodataset.org/}.

\end{itemize}

The baseline models, either image question generation model or image captioning model, take 5 images as input and generate 5 sentence, either 5 questions or 5 captions.
To align all settings, we concatenate 5 sentences from the baseline models and feed them into the same question generation model we use to generate questions for comparison.

% , either image question generation model or image captioning model, 
% To align with the VIST dataset setting, which the input is 5 images, we concatenate the 5 sentences of the baseline models, either image question generation model or image captioning model, into a short paragraph and take it as an input of question generation model.

\subsection{Human Evaluation}
We conduct human evaluation using crowd workers from Amazon Mechanical Turk (MTurk) to evaluate the quality of generated questions.
%, we conducted human evaluation through crowd-sourcing using the Amazon Mechanical Turk.
We randomly selected 250 photo sequences from the test set of VIST dataset and use three models (our model and two baselines) to generate questions for each photo sequence.
In the MTurk tasks, we show the photo sequence and three generated questions to the workers.
Each worker is asked ``which question is the best to start a conversation or keep the conversation continue'' and instructed to rank three questions.
We collect five response from five different workers for each photo sequence.

%Since there is no existing dataset or ground truth for this unique task, traditional n-gram-based automatic evaluation metric for language generation task, BLEU, METEOR score, etc., are not provided here.

% The user interface for annotators are shown in Figure \ref{fig:env}.
% \yun{Since the scenarios of datasets are different, we designed different instruction for annotators. For VIST dataset, ... For Chardes-Ego and Pepper-Life, ...}
% \begin{figure}[t]
%      \centering
%      \includegraphics[width=0.9\linewidth]{images/env.pdf}
%      \caption{The user interface, designed in a drag-and-drop fashion, for annotators to rank the questions. \yun{change instruction interesting}}
% \label{fig:env}
%  \end{figure}

%\subsection{User Study}
%\kenneth{Who were the participants? How were these participants recruited? What's the procedure of the study? Many details are missing here.}
%We also design questionnaires to ask each annotator about responsive question generation applications with robot and chatbot.
We also design questionnaires to explore potential directions of future work, in which
we ask annotators three questions: 
{\em (1)} If the chatmate is a \textbf{robot}, will you have a conversation with it? (Yes/No);
{\em (2)} Which one you would prefer if your chatmate is a \textbf{robot}? start a conversation yourself/wait for the robot to start the conversation; and
In Question {\em (3)} we ask again the same questions as (2), but change the chatmate to \textbf{your friend}.

\section{Results and Analysis}

\begin{figure}[t]
     \centering
     \includegraphics[width=\linewidth]{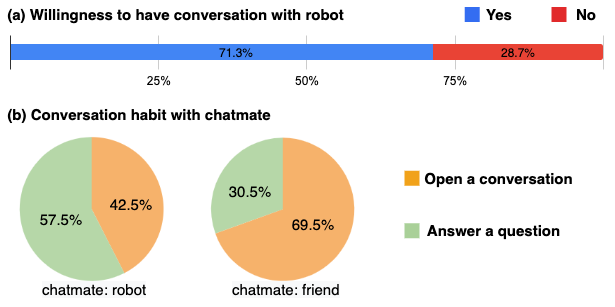}
     \caption{The result of survey question (1) that asks (a) ``Will annotators have conversation with robots?'', and questions(2) \& (3) that ask (b) ``Which conversation habit annotators prefer? with robot or friends as chatmate''.}
\label{fig:survey}
\end{figure}

\paragraph{User Survey}
The results of the designed questionnaires are shown in Figure~\ref{fig:survey}.
Among 3,750 annotators, exceeding 70\% annotators accept having a conversation with a robot, which validates chatbots' validity.
Figure~\ref{fig:survey}(b) shows the result of the questionnaire that asks users' conversation habits.
If the chatmates are annotators' friends, approximately 70\% of them tend to open a conversation actively.
However, if their chatmates are robots, they would wait for the robots to start a conversation and answer a question. 
This result shows the importance of the questions' quality if the chatbot demands to open a conversation with a human.

\begin{figure*}[t]
     \centering
     \includegraphics[width=0.97\linewidth]{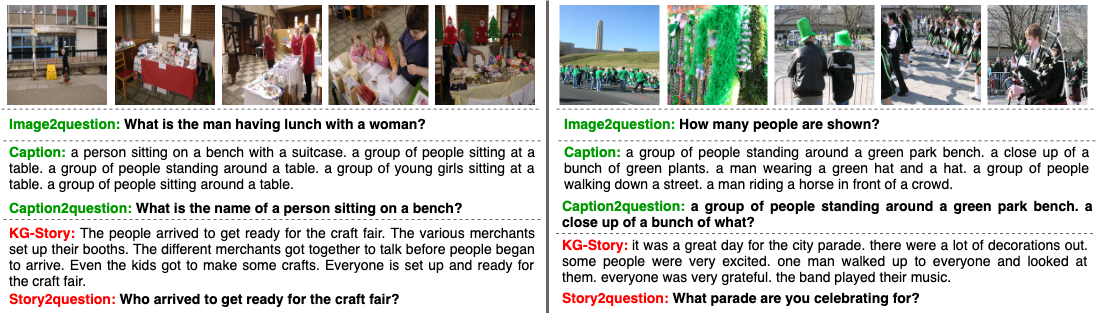}
     \caption{Examples of questions generated by different methods. }
\label{fig:ex}
 \end{figure*}
 
\paragraph{Quantity Analysis}
In Table~\ref{tb:human_ranking}, we show the rank of the questions generated from images (Img2Q), from captions (Caption2Q), and from stories (Story2Q).
Questions generated by three methods are ranked from 1 to 3 (lower is better).
Story2Q receives the best rank among baselines with the highest percentage (44.9\%) in rank 1 and the lowest percentage in rank 3 (23.2\%).
The fact that both Story2Q and Caption2Q outperform Img2Q shows that generating intermediate ingredients can have better understanding of images for generating questions than directly using image features as input.
In summary, Story2Q performs the best in generating response-provoking questions. 

% Caption2Q wins the second prize. Img2Q is the worst.
% Caption2Q gets the middle percentage among three methods in rank 1, rank 3, the highest percentage in rank 2 and the middle average rank. 
% Img2Q receives lowest percentage in rank 1, rank 2 and highest percentage in rank 3, and the highest average rank.
%All in all, story2Q performs the best in generating responsive questions by statistics.

In Table ~\ref{tb:question_dis}, we also show the composition of questions generated by each method by counting the 5W1H.
We find that Story2Q generates more diverse questions compared to Caption2Q and Img2Q.
Story2Q tends to generate various questions beginning with ``What", ``Why", ``When", ``Where", while Img2Q and Caption2Q mostly generate the questions begin with ``What".

% While img2Q and caption2Q generates more questions beginning with "what". 
% Last but not least, story2Q generates more diverse questions compared to caption2Q and img2Q. 
% In Table ~\ref{tb:question_dis}, we count the composition of questions of each method.
% The distribution of Story2Q's question is more balanced compared to img2Q and caption2Q.
% Story tends to generate various question beginning with "Where", "When", "Why",. While img2Q and caption2Q generates more questions beginning with "what". 
% \yun{We also count the composition of questions of each method.}

\begin{table}
\begin{center}
\small
\scalebox{0.95}{

\begin{tabular}{lcccc}
\hline
 Method & 1st & 2nd & 3rd  & Avg rank \\
\hline\hline

 img2Q  & 19.4\%(243) &  30.6\% & 49.9\%(624)  & 2.30 \\
 caption2Q  & 35.7\%(447) & 37.2\% & 26.9\%(337) & 1.91 \\
 {story2Q}  & \textbf{44.9\%}(\textbf{560}) & 32.2\% & \textbf{23.2\%(289)}  & \textbf{1.78} \\
\hline

%  & img2Q  & \%() & \% & \%  & \\
% Charades-Ego & caption2Q  & \%() & \% & \% &  \\
%  & {story2Q}  & \textbf{\%}(\textbf{}) & \% & \%  & \textbf{}  \\
% \hline

%  & img2Q  & \%() & \% & \%  & \\
% Pepper-Life & caption2Q  & \%() & \% & \% &  \\
%  & {story2Q}  & \textbf{\%}(\textbf{}) & \% & \%  & \textbf{}  \\
% \hline

\end{tabular}}
\end{center}
\caption{Human ranking evaluation between our proposed framework and two methods. 
First three columns indicate percentage of workers' ranking for each method, and last column denotes average rank (1 to 3, lower is better).
Numbers in brackets indicates the quantity of the best and the worst stories for each method.
The questions generated by our proposed framework is significantly better than all baseline methods with
$\rho < 0.05$.}
\label{tb:human_ranking}
\end{table}

\begin{table}
\begin{center}
\small
\scalebox{0.83}{

\begin{tabular}{lccccccc}
\hline
 Method & What & Where & When  & Why & Who & How & Other \\
\hline\hline

 img2Q  & 70.7\% & 1.4\% & 5.1\% & 3.0 \% & 0\% & 13.8\% & 6.0\%   \\
 caption2Q  & 79.2\% & 1.1\% & 2.7\% & 3.7\% & 0\% & 7.6\% & 5.7\%   \\
 {story2Q}  & 55.5\% & 5.6\% & 11.6\% & 15.3\% & 1.0\% & 6.3\% & 4.7\%  \\
\hline

\end{tabular}}
\end{center}
\caption{The question distribution of different methods.}
\label{tb:question_dis}
\end{table}

\paragraph{Quality Analysis}
Img2Q prefers to generate literal questions that ask the quantity and information in the images, which is not suitable for starting a conversation.
Take the right half of Figure ~\ref{fig:ex} as an example, Img2Q asks ``How many people are shown?", which is considered a literal question that can not make the conversation lasting. %start a conversation
Img2Q directly uses features extracted from images to generate questions. The failure of this end-to-end method lets us consider developing a multi-stage model and using extra methods like object detection and relation extraction to support engaging question generation.

Since captioning is good at summarizing all the contents in an image with the attention mechanism's help, the information in the question strongly matches with the image sequence.
However, the questions generated by Caption2Q lack creativity and diversity, since captions are only designed to be faithful to the original images.
In the left example of Figure~\ref{fig:ex}, Caption2Q asks ``What is the name of a person sitting on a bench?", since all of the contents can be easily found in the image sequence, the question may be considered boring and obvious by human.  

Fortunately, Story2Q generates response-provoking questions with high quality because it firstly generates interesting stories by object detection and relation extraction. 
Since the story contains deeper relations among different objects in the image sequence, the generated question is profound. 
In Figure 4, Story2Q asks ``Who arrived to get ready for the craft fair?". 
It takes care of the ``arrive" relation between human and craft fair.
Paying attention to relationships among different objects is similar to human thinking, so it is more natural and reasonable to start a conversation with this kind of question.
%which is natural and reasonable to start a conversation. 
%While other two methods only care about the name and the information about the person. Human may think questions generated by story2Q is more interesting and diverse, and it's easier for human to start a conversation.

 %Story2Q firstly generates an imaginative story, the content in the image may lose a little bit to improve diversity and interest of the story. So when the story is transformed to question, some contents in the image miss compared to question generated by cap2Q.  

%dataset are different, VIST is more ... than Charades-Ego and Pepper-Life......
%Take Fig.~\ref{fig:ex} as example, the questions generated based on stories are more interesting.

%put 3 big figures and comparison (of dataset or question quality) analysis here

\begin{figure}[t]
     \centering
     \includegraphics[width=\linewidth]{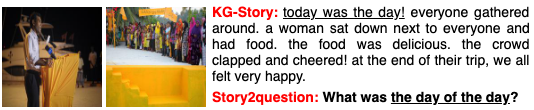}
     \caption{Repetition error example of question generation model.}
\label{fig:error}
 \end{figure}

\paragraph{Error Analysis}
Since the story generation model and question generation model are trained on different datasets({\em e.g.}, VIST and SQuAD), we find that question generation model tends to copy the same sentence or predict repetitive words when meeting the sentence it did not learn before.
Taking the question generated from a caption in the right example of Figure~\ref{fig:ex} as an example, the story generation model directly copies the whole sentence of the caption and generates a strange question.
In Figure~\ref{fig:error}, KG-Story sometimes generates sentence that beyond images, like ``today was the day!" for this example.
This kind of sentences is hard to exist in the SQuAD dataset, thus resulting in the chaos of the question generation model and yielding the model to generate repetitive words.

\section{Conclusion and Future Work}

We have introduced a new scenario of visual question generation, in which, when given a sequence of images, the system should generate a provocative question in order to start a conversation.
Instead of generating question directly from image features, we ask the model imagine first by generating creative stories to have a better understating of an image sequence, and then produce an engaging question based on stories to start a conversation.
Human evaluation results show that our proposed framework significantly outperforms two baseline models, Img2Q and Cap2Q, on the VIST dataset.
This provides evidence that thinking before asking can enhance the question's quality and make people want to communicate with the system.

% imagine first so that it candgenerate an engaging question to start a conversation.
% The human evaluation result shows that our system outperforms the two baseline model, img2Q and caption2Q, on the VIST dataset. 
% This provides evidence that thinking before asking can enhance the question's quality and make people want to communicate with the system.

% future work
% use more dataset to evaluate?
% propose a new dataset
% ...
% robot ?
There are several potential future research directions. 
Our model is currently only evaluated on one vision-to-language dataset ({\em e.g.}, VIST), and thus we want to explore the generalization of our idea to other datasets.
% how to generate our idea to other image or robot-vision datasets.
Besides, according to the result of experiments and human feedback, we see this \vToq task's potential. 
We can collect a dataset, which will launch a new challenge to the community and further invoke interests in studying the importance of asking a response-provoking question.

\section{Acknowledgement}
This research is partially supported by {\em (i)} Ministry of Science and Technology, Taiwan under the project contract 108-2221-E-001-012-MY3 and 
{\em (ii)} College of IST Seed Fund at Penn State University.
We also thank the workers on MTurk who participated in our studies.

% Ministry of Science and Technology, Taiwan under the project contract 108-2221-E-001-012-MY3.

\bibliographystyle{aaai21}
\bibliography{aaai2021}
% \input{main.bbl}
% \bibliographystyle{aaai2021}
% \bibliography{aaai2021}

\end{document}